\documentclass[journal]{IEEEtran}
\usepackage{epsfig} 
\usepackage{amsmath} 
\usepackage{amssymb}  
\usepackage{graphicx}
\usepackage{epstopdf}
\usepackage{upgreek}
\usepackage{caption}
\usepackage{color}
\usepackage{subcaption}
\usepackage{dblfloatfix}
\usepackage[utf8]{inputenc}
\usepackage{tikz}
\usepackage{esint}
\usepackage{accents}

\graphicspath{ {./Images/} }

\DeclareRobustCommand{\groupderiv}[1]{\accentset{\scriptstyle\circ}{#1}}
\DeclareMathOperator*{\argmax}{arg\,max}

\definecolor{matred}{rgb}{0.85,0.325,0.098}
\definecolor{matpurple}{rgb}{0.494,0.184,0.556}

\newcommand{\vecA}{\mathbf{A}}

\newcommand{\vecC}{\mathbf{C}}
\newcommand{\vecD}{\mathbf{D}}

\newcommand{\vecF}{\mathbf{F}}
\newcommand{\vecG}{\mathbf{G}}
\newcommand{\vecH}{\mathbf{H}}

\newcommand{\vecJ}{\mathbf{J}}

\newcommand{\vecM}{\mathbf{M}}

\newcommand{\vecR}{\mathbf{R}}
\newcommand{\vecS}{\mathbf{S}}

\newcommand{\vecb}{\mathbf{b}}

\newcommand{\vecf}{\mathbf{f}}

\newcommand{\vecq}{\mathbf{q}}
\newcommand{\vecr}{\mathbf{r}}

\newcommand{\vecu}{\mathbf{u}}
\newcommand{\vecv}{\mathbf{v}}

\newcommand{\vecz}{\mathbf{z}}

\newcommand{\vecphi}{\boldsymbol\upphi}

\newcommand{\veclambda}{\boldsymbol\lambda}

\newcommand{\vecphid}{\dot{\boldsymbol\upphi}}

\newcommand{ \vctwo }[2] {
 \left[\begin{array}{c }
  #1        \\
  #2
\end{array} \right]  }

\newcommand{ \vcthree }[3] {
 \left[\begin{array}{c }
  #1        \\
 #2\\
  #3
\end{array} \right]  }

\newcommand{ \matwo }[4] {
 \left[\begin{array}{c c }
   #1  &  #2     \\
   #3  &  #4
\end{array} \right]  }

\newcommand{ \mathree }[9] {
 \left[\begin{array}{c c c}
   #1  &  #2  &  #3   \\
   #4  &  #5  &  #6   \\
   #7  &  #8  &  #9
\end{array} \right]  }

\title{Geometric analysis of gaits and optimal control for three-link kinematic swimmers}
\author{Oren Wiezel, Suresh Ramasamy, Nathan Justus, Yizhar Or and Ross Hatton
\thanks{The work of OW and YO has been supported in part by Israel Ministry of Science and Technology, under contracts no. 3-15622 and  3-17383}
\thanks{The work of SR, NJ and RH has been supported in part by the National Science Foundation awards 1653220 and 1935324}
\thanks{OW and YO are with the Faculty of Mechanical Engineering at the Technion - Israel Institute of Technology, Haifa 3200003, Israel (e-mail: orenwi@campus.technion.ac.il, izi@technion.ac.il). }
\thanks{SR, NJ and RH are Oregon State University, Corvallis, OR 97333 USA. (e-mail: ramasams@oregonstate.edu , justusn@oregonstate.edu, Ross.Hatton@oregonstate.edu).}}

\begin{document}

\maketitle

 \begin{abstract}
 Many robotic systems locomote using gaits -- periodic changes of internal shape, whose mechanical interaction with the robot's environment generate characteristic net displacements. Prominent examples with two shape variables are the low Reynolds number 3-link ``Purcell swimmer'' with inputs of 2 joint angles and the ``ideal fluid'' swimmer. Gait analysis of these systems allows for intelligent decisions to be made about the swimmer's locomotive properties, increasing the potential for robotic autonomy.
In this work, we present comparative analysis of gait optimization using two different methods.
The first method is variational approach of ``Pontryagin's maximum principle'' (\textit{PMP}  ) from optimal control theory. We apply \textit{PMP}   for several variants of 3-link swimmers, with and without incorporation of bounds on joint angles.
The second method is differential-geometric analysis of the gaits based on curvature (total Lie bracket) of the local connection for 3-link swimmers.
Using optimized body-motion coordinates, contour plots of the curvature in shape space give visualization that enables identifying distance-optimal gaits as zero level sets.
Combining and comparing results of the two methods enables better understanding of changes in existence, shape and topology of distance-optimal gait trajectories, depending on the swimmers' parameters.
 \end{abstract}

\section{Introduction}
Robotic swimmers are a promising avenue of research. Both small microswimmers and large scale swimmers have many promising possible applications.
Advances in technology for manufacturing nano- and micro-systems have brought renewed interest in simplified models of microswimmers and production of microscopic robotic devices that could be applied in the medical field \cite{gao2012cargo,cho2014mini,kosa2008flagellar}.
Such micro-robots would be able to provide targeted drug delivery, tumor detection, assisted sperm motility, and even perform minimally invasive surgical procedures. At the other end of { size} scale, large, fast moving swimming robots can be useful in search and rescue missions, maintenance operations
within pipe systems of complex infrastructures, and surveillance
or protection in marine environments \cite{Tur2010Robotic}.

The flow around microswimmers is governed by Stokes equations, which arise from the Navier-Stokes equations at the limit of zero Reynolds number $Re \to 0$ \cite{happel&brenner_book,CohenSwimming2010}. At very small Reynolds numbers, $Re\ll 1$, inertial forces become negligible and the viscous forces are dominant. This leads to a linear relationship between the body velocities and the internal shape velocities \cite{koiller1996problems}.
These unique attributes call for drastically different swimming strategies than the ones used in the familiar motion of large organisms.

\begin{figure}[t!]
\centering
\begin{subfigure}[c]{0.24\textwidth}
\includegraphics[width=\textwidth]{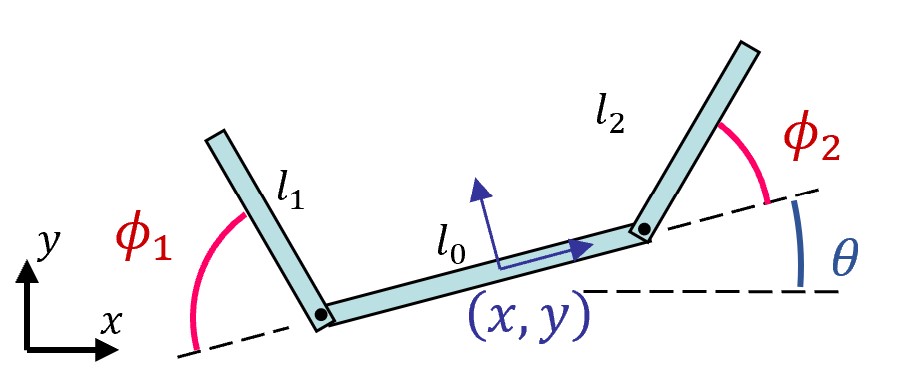}
\caption{}
\label{fig:Purcell_swimmer}
\end{subfigure}
\begin{subfigure}[c]{0.24\textwidth}
\includegraphics[width=\textwidth]{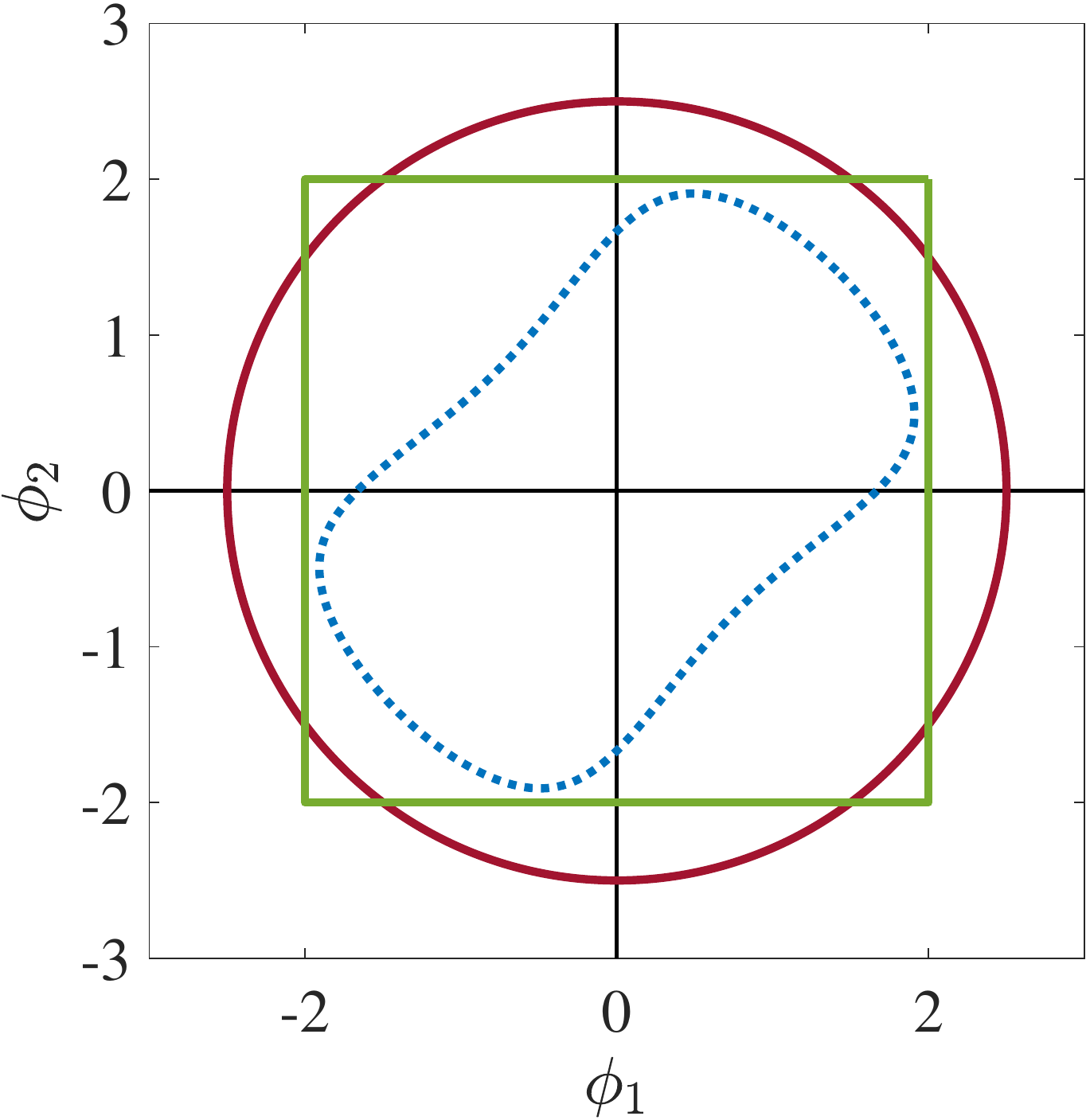}
\caption{}
\label{fig:sq_cir_Tam}
\end{subfigure}
 \caption{(a) Purcell's three-link swimmer. (b) The square, circular and maximal displacement gaits.}
 \label{Fig:Purcell_model&gaits}
 \end{figure}

\begin{figure}[t!]
 \centering
 \begin{tikzpicture}
    \node[anchor=south west,inner sep=0] (image) at (0,0,0) {\includegraphics[width=0.3\textwidth]{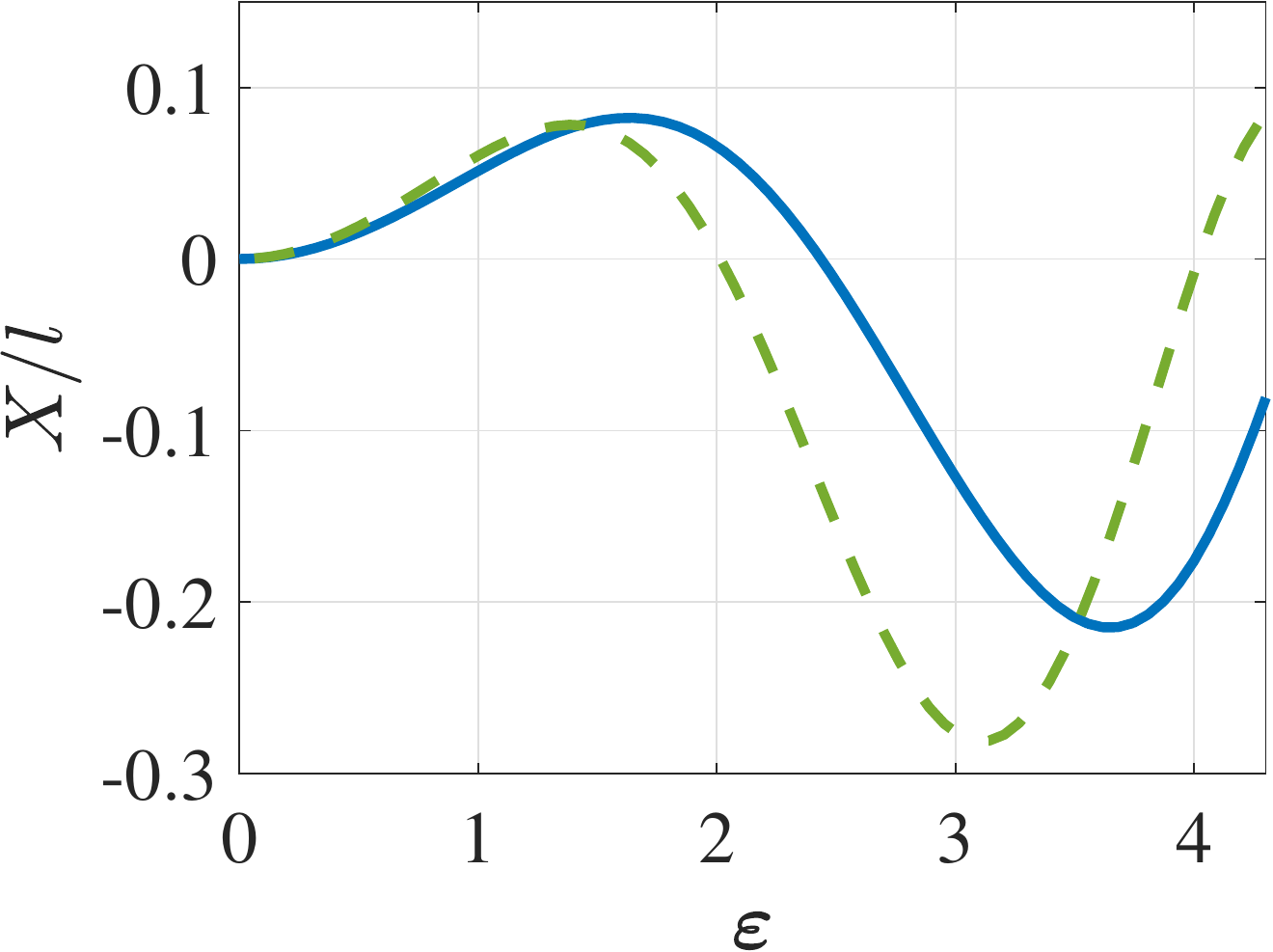}};
    \begin{scope}[x={(image.south east)},y={(image.north west)}]
         \draw (0.5,0.875) node[matpurple,rotate=45]{\huge +};
        \draw (0.87,0.345) node[matred,rotate=45]{\huge +};
    \end{scope}
\end{tikzpicture}
\caption{Net-displacement over a period in the $x$ direction as a function of the gait amplitude $\varepsilon$. Circular gait in blue full line and square gait in dashed' green line. X marks denote optimal amplitudes for maximal (purple) and minimal (red) displacement for circular gait.}
 \label{fig:Purcell_disp}
 \end{figure}

Purcell suggested that the ``simplest animal'' that could swim is the three-link swimmer \cite{purcell1977life}, comprised of three thin rigid links connected by two rotary joints (Fig. \ref{fig:Purcell_swimmer}). By alternately rotating the joint angles, this swimmer would be able to propel itself in a low-$Re$ environment.
The series of shape changes, or ``gait'' proposed by Purcell appears as a square in the plane of shape variables (the joint angles) as shown in Fig. \ref{fig:sq_cir_Tam}.
This gait results in motion along the mean orientation of the central link. Becker \textit{et al.} \cite{becker2003self} formulated the dynamics of Purcell's swimmer using \textit{slender-body theory} \cite{cox1970,batchelor1970slender} and found that the direction of net displacement of the swimmer depends on the angular amplitude of the strokes.
For small amplitude the swimmer will move in one direction, but for larger amplitudes the translation will be in the opposite direction (Fig. \ref{fig:Purcell_disp}).
This also implies the existence of an amplitude of the gait that will result in maximal translation (marked with a Purple X in Fig. \ref{fig:Purcell_disp}). For very large amplitudes there is a second maximum-displacement gait that will result in swimming in the opposite direction (marked with a red X in Fig. \ref{fig:Purcell_disp}).\footnote{We note that maximizing displacement per cycle is not a true ``optimal control" problem for the systems considered in this paper, because per-cycle displacement does not account for the different amounts of time or energy consumed by ``large" or ``small" gaits.  Displacement-maximizing gaits do, however, have a distinct mathematical structure and exist in close proximity to speed and efficiency maximizing gaits \cite{hatton2017cartography}, and so are therefore an interesting topic of study.} { Importantly, in robotic realizations of such swimmers, cf. \cite{Gutman2015Symmetries}, one commonly has to account for practical bounds on joint angles which may limit the stroke amplitude.}

In contrast to small microswimmers whose motion is governed by viscous drag forces, the motion of large swimmers, such as fish or eels, is governed mainly by inertial forces \cite{KMlocomotion,melli2006motion}. A common simplified mathematical model for such swimmers is that of ``perfect fluid'' that assumes potential flow in a inviscid fluid and allows for an \textit{added mass effect} \cite{lamb1945hydrodynamics}. Notably, using this model for a swimmer in an unbounded fluid domain that dictates symmetry and conservation laws leads to first order equations of motion that give a linear relation between the body velocities and internal shape velocities. Therefore, motion analysis of an inertial swimmer under this ``perfect fluid'' model can be done using the same mathematical methods as the in the case of a microswimmer, despite the  differences in the physics of the mechanism creating the motion \cite{KMlocomotion,morgansen2007geometric}. Previous works have examined the reduction of inertia-driven ``perfect fluid'' dynamics into a first order system \cite{KMlocomotion,virozub2019planar,hatton2013tro}.

The common basis for the small-scale microswimmers and large-scale ``perfect fluid''  swimmers is that their motion can be described by a ``principal kinematic'' structure \cite{hatton2013tro}, where the body velocities are linearly related to the shape velocities. { Based on the Lie-group structure of rigid-body motion, } this locomotion principle is not unique to swimming and is shared by other models of robotic locomotion induced by internal shape changes, such as the terrestrial snake \cite{Marvi224} and some wheeled robots \cite{yona2019wheeled,shammas2007geometric}.

Optimization of design and actuation of low-$Re$ swimmers has been a subject of several recent studies. One approach to gait optimization was taken in \cite{hatton2011geometric}, where geometric mechanics techniques based on Lie brackets were used to provide an approximation of the net displacement for locomoting systems.
These approximations are then used to search for the optimal gait. Historically, the approximation error has limited this analysis to small shape changes.
Recent works by Hatton and Choset \cite{hatton2015epj} show that the choice of body coordinates can significantly reduce the approximation error. This allowed them to find gaits maximizing displacement per cycle of the three-link swimmer by examining zero-contour lines of the total Lie bracket, as well as gaits for maximal energetic efficiency which also maximize displacement per time at a given effort level \cite{ramasamy2019geometry}.

Another { important} approach for finding optimal gait trajectories is using variational methods of optimal control, namely Pontryagin's maximum principle (\textit{PMP}  ) {\cite{pontryagin1987mathematical}, \cite{bryson1975applied}, {\cite{Liberzon2011book,clarke2013functional}. This approach has been vastly applied in several practical fields such as aerospace applications \cite{ben2010optimal,serra2021tracklet,niknam2022optimal} and computer vision \cite{kimia1994optimal}, as well as biology \cite{rosa2018optimal,sharp2021implementation} and quantum dynamics \cite{bonnard2012optimal,boscain2021introduction}}. For Purcell's three-link microswimmer model, the \textit{PMP} method} has been applied in \cite{wiezel2016using}, and analytically reproduced the maximum displacement-per-cycle gait found in previous works \cite{tam2007optimal,hatton2013tro} using different approaches. The gait in \cite{wiezel2016using} was found using a polar representation of the shape variables and so is not suitable for closed curves with a general shape. Another limitation of the  \textit{PMP}   formulation in \cite{wiezel2016using} is that it did not allow for bounds on the shape variables, which commonly exist in any practical application.

{ Extension of the maximum principle to account for inequality constraints on state variables have been presented in several works, e.g. \cite{ben2010optimal,chang1961optimal,dubovitskii1968necessary,Speyer_Bryson_boundedstate,Hartl_survey_boundedstate}. This extension has also been recently incorporated in optimal control for specific applications such as autonomous transportation and vehicles \cite{ritzmann2019fuel,zhou2019state,ben2021time}, energy management systems \cite{nguyen2021multi,lee2021optimal}, and aerospace systems \cite{cots2018direct,mall2020unified}.

A different variant of optimal control, which has significance for computational aspects, is the {\em discrete-time PMP} \cite{holtzman1966convexity,bolt1975method,dubovitskii1978discrete}. In particular, several works have focused on systems evolving on Lie groups, and development of discrete integrators that preserve group invariants and thus avoid cumulative integration error \cite{kobilarov2011discrete,saccon2013optimal,phogat2018discrete}. This method can incorporate bounds on input and state variables due to its discrete nature. For Purcell's three-link microswimmer model, the recent work \cite{kadam2021exact} considered the {\em iso-holonomic problem} of finding gaits that minimize energy expenditure under given net displacement per cycle under given state bounds, by employing discrete-time {\em PMP} on a Lie group.

The goal of this work is to revisit the kinematic models of three-link swimmers and explore optimal gait trajectories of joint angle inputs} for achieving maximal net displacement per cycle, { with possible bounds on joint angles. We make a comparative analysis of two different methods. The first method is based on necessary conditions of variational calculus, using {\em PMP} within the framework of optimal control. It requires solving a two-point boundary-value problem which reduces to a simple ``shooting'' solution of a scalar variable. The method captures locally-optimal smooth gait trajectories for maximal positive displacement. However, for some swimmer’s model parameter it fails to find a smooth trajectory of large-amplitude gait for maximizing negative displacement, whereas a non-smooth locally-optimal gait can be found when state bounds are incorporated.

{The second method is based on differential geometry and on using an approximate solution (formally a truncated Baker-Campbell-Hausdorff series) to the net-displacement-per-cycle equations. This approximation allows the net displacement over a gait cycle to be computed as the volume integral of a scalar height function over the region enclosed by the gait trajectory. Through a principled choice of coordinates for the system (in particular, selecting a body frame at a nonlinearly weighted average of the link positions instead of being attached to a link), we have previously shown that this approximation can be made usefully accurate over large-amplitude motions of the joints~\cite{hatton2011geometric}.

In particular, a very accurate approximation of gait trajectories that maximize displacement per cycle is obtained by the loops of zero-level curves of the scalar height function. Upon varying swimmer's parameters, changes in height function landscape and in topology of its zero-level curves give a graphical explanation to the change in shape and nature of such gait trajectories. In particular, junction points in the zero-level curves preclude existence of  locally-optimal trajectories, and state bounds become necessary in order to obtain non-smooth optimal gaits that avoid the junction points. Overall, the two methods reconcile together to combine a rigorous variational formulation with graphical visualization of the changes in shape and nature of optimal gaits.
}

The structure of the paper is as follows. In the next section, we present two models of swimmers: Purcell's swimmer and the ``perfect fluid'' swimmer.
In section \ref{sec:OCP} we offer the formulation of the optimal control problem using \textit{PMP}, and show our solutions for the models {with and without joint angle bounds,} as well as cases where this formulation fails to find { smooth} optimal gaits.
Next, in section \ref{sec:GAG} we review the geometric analysis of gaits introduced in \cite{hatton2011geometric}. 
In section \ref{sec:compare} we compare the results of the two methods. 
{ The closing section contains discussion and conclusion.}

\section{Mathematical models of three-link kinematic swimmers}
\label{sec:models}
We begin by formulating the dynamics of two models of 3-link swimmers. The swimmer model consists of three thin rigid links with lengths $l_0,l_1,l_2$, where $l_1=l_2$ . The links are connected by two rotary joints whose angles are denoted by $\phi_1$ and $\phi_2$ (see Fig. \ref{fig:Purcell_swimmer}). The shape of the swimmer is denoted by the vector $\vecphi=(\phi_1,\phi_2)^T$. It is assumed that the swimmer's motion is confined to a plane. The planar position of the middle link's center is $(x,y)$ and its orientation angle is $\theta$. The velocity of the central link in an inertial frame is denoted by $\dot \vecq_b=(\dot x, \dot y, \dot\theta)$. The velocity of the $i$th link is described by the linear velocity of its center and the links angular velocity $\omega_i$, which are augmented in the vector $\vecv_i=(\dot x_i,\dot y_i,\omega_i)\in\mathbb{R}^3$.
We denote the body velocities by $\mathring{\vecq}=(v_x, v_y, \dot\theta)$. These are the velocities of the central link expressed in a reference frame attached to the central link. We show below that for each of the models, the relation between the shape velocities and the body velocities can be written as:
\begin{equation}
\groupderiv{\vecq}=\vecA(\vecphi)\dot\vecphi
\label{eq:body_vel_connection}
\end{equation}
also known as the kinematic reconstruction equation \cite{bloch2003nonholonomic}.
In order to obtain the velocities in an inertial frame we multiply by a rotation matrix:
\begin{equation}
\dot\vecq_b=\vecR(\theta)\groupderiv{\vecq}=\vecR(\theta)\vecA(\vecphi)\dot\vecphi=\vecG(\theta,\vecphi)\dot\vecphi
\label{eq:full_connection}
\end{equation}
While in this work we only consider planar three-link models, this form can also be generalized to multi-link models and to spatial swimmers using three rotation angles \cite{HelicalOrNelsonChapnik2021}.

In the rest of this section, we give a short review on how to write the relation \eqref{eq:body_vel_connection} for both swimmer models.

\subsection{Purcell's swimmer}
{The first swimmer model we consider is ``Purcell's swimmer", in which the system is submerged in an unbounded fluid domain whose motion is governed by Stokes equations \cite{happel&brenner_book}.
Here, we briefly summarize how to find the local connection for Purcell's swimmer.  More in-depth derivations can be found using inertial-frame and link-frame methods described in \cite{ramasamy2019geometry} and \cite{wiezel2016optimization}.}

First, the velocity of each link through the fluid is written using Jacobian { matrices $\vecJ_i$} relating overall swimmer body velocity and joint velocities to individual link velocities:

\begin{equation}
    \vecv_i = \vecJ_i\begin{bmatrix}\mathring{\vecq}\\\dot\vecphi\end{bmatrix}
    \label{eq:jacobian}
\end{equation}

Resistive force theory is used to calculate the force on each link as being linearly proportional to that link's velocity { $\vecv_i$ through a  resistance matrix $\vecD_i$ }:

\begin{equation}
    \vecF_i = -\vecD_i\vecv_i
\end{equation}

Then, forces on each link can be mapped back into the body frame of the swimmer using a dual adjoint-inverse mapping and summed together { to obtain net load on the body $\vecF_b$:}

\begin{equation}
    \vecF_b = \sum_{i=0}^2{\vecJ_i^T}\vecF_i = \left(\sum_{i=0}^2-{\vecJ_i^T}\vecD_i\vecJ_i\right)\begin{bmatrix}\mathring{\vecq}\\\dot\vecphi\end{bmatrix}
\label{eq:Fb0} \end{equation}

At low Reynolds numbers, net forces on the swimmer are zero because it is at quasistatic equilibrium.  This allows us to write { equation \eqref{eq:Fb0} as nonholonomic constraints on velocities:}

\begin{equation}
    \vecF_b = \vecC_\vecq{(\vecphi)}  \mathring{\vecq} + \vecC_{\vecphi}{(\vecphi)}  \dot\vecphi = 0
\end{equation}

This allows us to solve for the viscous local connection:

\begin{equation}
    \mathring{\vecq} = -\vecC_\vecq^{-1}\vecC_{\vecphi}\dot\vecphi = \vecA(\vecphi)\dot\vecphi
\end{equation}


\subsection{Perfect fluid swimmer}
The second swimmer we consider is the ``perfect fluid'' swimmer model (Fig. \ref{fig:3-link_perfect}), whose dynamics are constructed by assuming inviscid and irrotational potential flow.
In \cite{virozub2019planar}, analysis for sinusoidal input was done and motion experiments were performed with a controlled robotic swimmer in order to validate this model.
Each link is assumed to be an ellipse with principal radii of $a_i$, $b_i$ and density $\rho$, which has mass $m_i$ and moment of inertia $I_i$. The swimmer is submerged in an unbounded domain of ideal fluid with equal density $\rho$.


The kinetic energy of the system can be written as a sum of the kinetic energies of the individual links:
\begin{equation}
T=\frac{1}{2}\sum_i\vecv_i^T\vecM_i\vecv_i
\label{eq:kinetic}
\end{equation}
where
$$
\vecM_i=\mathree{m_i}{0}{0}{0}{m_i}{0}{0}{0}{I_i}+\pi\rho\mathree{b_i^2}{0}{0}{0}{a_i^2}{0}{0}{0}{\frac{1}{8}(a_i^2-b_i^2)}
$$
The first term in $\vecM_i$ represents the inertia of the link while the second term represents the added mass effect due to accelerating the displaced fluid around the swimmer \cite{KMlocomotion}.

Using the link Jacobian to relate body-frame velocity and joint velocity to individual link velocity as in (\ref{eq:jacobian}), we can rewrite the kinetic energy in terms of these two quantities:
\begin{equation} \label{eq:T_M_blocks}
T=\frac{1}{2}\vctwo{\mathring{\vecq}}{\vecphid}^T\matwo{\vecM_{bb}}{\vecM_{bs}}{\vecM_{bs}^T}{\vecM_{ss}}\vctwo{\mathring{\vecq}}{\vecphid}
\end{equation}
where all blocks $\vecM_{bb},\vecM_{bs},\vecM_{ss}$ are functions of the shape variables $\vecphi$ only.

\begin{figure}[t]
\centering
\includegraphics[width=0.4\textwidth]{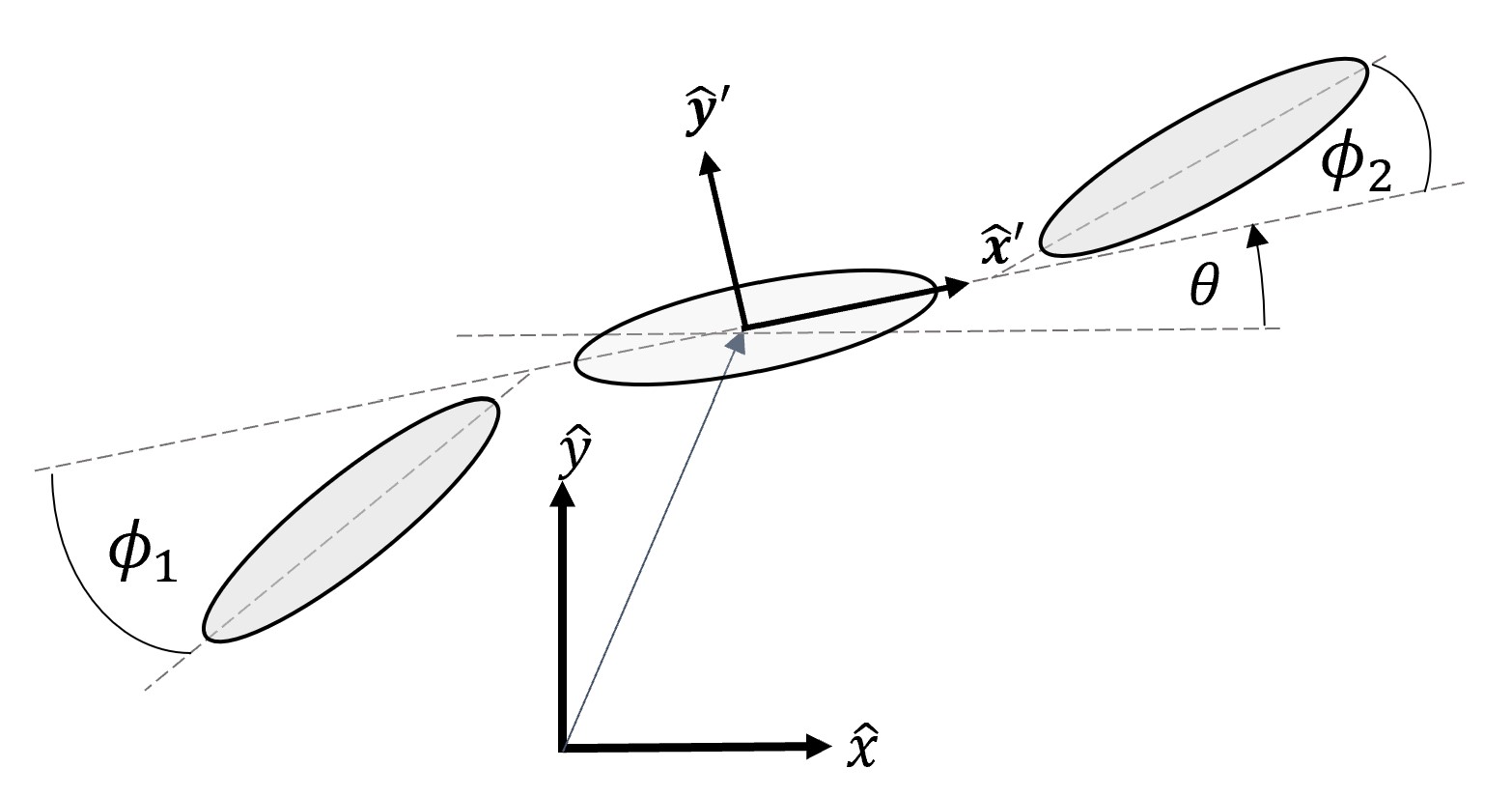}
\caption{The three-link ``perfect fluid'' swimmer model \cite{virozub2019planar}.}
 \label{fig:3-link_perfect}
 \end{figure}

{ Using generalized coordinates $\vecq=(\vecq_b,\vecphi)$, the system's dynamic equations of motion can be written using Lagrange's formulation \cite{hand1998analytical} as
\begin{equation}
\frac{d}{dt}\left(\frac{\partial T}{\partial \dot \vecq}\right)-\frac{\partial T}{\partial \vecq}=\vecF_q ,
 \label{eq:Lagrange}
 \end{equation}
where $\vecF_q$ is a vector of generalized forces. In matrix form, this equation reads as
\begin{equation}
\vecM(\vecq)\ddot{\vecq}+\vecb(\vecq,\dot \vecq)=\vecF_q . \label{eq:MBFq}
 \end{equation}
The system's inertia matrix $\vecM(\vecq)$ in \eqref{eq:MBFq} has tight relation to the block matrices in \eqref{eq:T_M_blocks} due to invariance of the dynamics with respect to rigid body transformations, which induces conservation of generalized momentum variables, see \cite{virozub2019planar,hatton2013tro,shammas2007geometric}. Assuming that the swimmer starts from rest, $\vecq(0)=0$, this momentum conservation can be reduced to a first-order}
%
%
relation between body velocity and shape changes as (see \cite{KMlocomotion}):
\begin{equation}
\vecM_{bb}(\vecphi)\mathring{\vecq}+\vecM_{bs}(\vecphi)\dot\vecphi=0
\end{equation}
The body velocities can now be written in the form of equation \eqref{eq:body_vel_connection} with $\vecA(\vecphi)=-\vecM_{bb}^{-1}\vecM_{bs}$.


\section{Optimal control using PMP}
\label{sec:OCP}
{ Applying periodic input of shape changes (gaits) to the swimmer models presented above, results in net motion per cycle, which can be obtained by integrating \eqref{eq:full_connection} over a period.
Importantly, the kinematic relation \eqref{eq:full_connection} is time-invariant, and corresponds to a geometric relation
\begin{equation}
d \vecq_b=\vecG(\theta,\vecphi) d \vecphi .
\label{eq:geom_connection}
\end{equation}
This implies that for any geometric shape of gait trajectory $\vecphi(s)$, the resulting net body motion $\Delta \vecq_b$ generated by this gait is independent of rescaling the time-rate $s(t)$ of the input \cite{kelly1995geometric}. One may seek for optimal gaits that maximize per-cycle displacement \cite{hatton2013tro}.
}
In order to find gaits that maximize per-cycle displacement of the swimmer models presented, we apply methods of variational optimal control, which lead to the \textit{PMP}  \cite{bryson1975applied}.
We first present the formulation of a general solution via \textit{PMP}   and the formulation for our problems with dynamics \eqref{eq:body_vel_connection}, with and without state bounds.
Next, we present the optimal gaits found using this method for both of the swimmer models and the influence of changes in problem's bounds and model parameters on the optimal solution.

\subsection{Formulation of OCP}
We start with a short description of the optimal control problem and \textit{PMP}   solution for a model with the dynamics in \eqref{eq:body_vel_connection}.

For an optimal control problem
\begin{equation}
(OCP)
	\begin{cases}
	\max J=\int_0^{t_f} g(\vecz,\vecu) \ s.t.\\
	\dot\vecz=\vecf(\vecz,\vecu)& \forall t\in [0,t_f],\\
	\vecu \in \mathcal{U} & \forall t \in [0,t_f],\\
	\vecz(0)=\vecz_0, &
	\end{cases}
\end{equation}
where $\vecz\in\mathcal{R}^n$ is the state of the system and $\vecu$ is the control input, we define a costate vector $\veclambda\in\mathcal{R}^n$ and the Hamiltonian will be
\begin{equation}
H(\vecz,\vecu,\veclambda)=\veclambda^T \vecf+g
\end{equation} 
 Pontryagin's Maximum Principle (\textit{PMP}) states that { a necessary condition for obtaining} the optimal control input trajectory $\vecu^*(t)$ with associated state trajectory $\vecz^*(t)$ is { that it} maximizes the Hamiltonian:
 \begin{equation}
 \vecu^*(\vecz,\veclambda)=\argmax_{\vecu\in\mathcal{U}} H(\vecz,\vecu,\veclambda).
 \end{equation}
{ Thus, an } optimal input can be found by solving
 \begin{equation}
 \vecH_\vecu=\frac{\partial H}{\partial \vecu}=\vec{0}
 \label{eq:Hu_general}
 \end{equation}.

The Maximum Principle requires solving an ODE with two-point boundary conditions.

\subsubsection*{Singular arcs}
In problems where the Hamiltonian is linear in the input, the optimal input cannot be found from \eqref{eq:Hu_general}. In many cases, this implies a ``bang-bang'' solution where the control switches between the upper and lower bounds. The switching is determined by the sign of $\vecH_\vecu$.
In some cases, when $\vecH_\vecu$ can vanish for a finite time, the solution follows a singular arc and can be determined by the time derivatives:
\begin{equation}
 \frac{d^k}{dt^k}\vecH_\vecu=\vec{0}
 \label{Hu_time_der}
 \end{equation}
After an even number of derivatives $k$, the input $\vecu$ appears and can be extracted.
\subsubsection*{Bounded state}

When formulating the maximum principle with inequality bounds on the state variables, we follow the \textit{direct adjoining approach} \cite{Speyer_Bryson_boundedstate,Hartl_survey_boundedstate}.
For an inequality bound in the form:
\begin{equation} \label{eq:bound}
w(\vecz,t)\leq 0
\end{equation}
with time derivatives
\begin{equation}
w^k(\vecz,t)=\frac{d^k w}{dt^k}
\end{equation}
The Hamiltonian is defined as
\begin{equation}
\tilde H(\vecz,\vecu,\veclambda,\nu)=\veclambda^T \vecf+g+\nu w
\end{equation}
where $\nu$ is an additional multiplier, and
\begin{equation}
\nu w^*(\vecz,t)=0
\end{equation}
so that $\nu=0$ whenever the bound is inactive ($w(\vecz,t)<0$), { while} $w(\vecz,t)=0$ { and $\nu>0$} whenever the bound is active.

A time interval $[\tau_1, \tau_2]$ is called a \textit{bounded interval} if $w(\vecz,t)=0$ for $t\in [\tau_1, \tau_2]$ (the bound is active).
$\tau_1$ and $\tau_2$ are the entry time and exit time, respectively.

Whenever the bound is inactive ($w(\vecz,t)<0$), the solution is found through $\vecH_\vecu=\vec{0}$ (or its time derivatives \eqref{Hu_time_der} in the case of a singular arc). When the bound is active, $w(\vecz,t)=0$ and its time derivatives determine the control input $u^*$.
Generally, the costate variables as well as the Hamiltonian do not necessarily change continuously at entry time and exit time and additional necessary conditions can be written that determine the discontinuity \cite{Speyer_Bryson_boundedstate}. But, if the control input is discontinuous over the entry or exit time, $u^*(\tau^-)\neq u^*(\tau^+)$, then the costate variables and Hamiltonian must be continuous, $\veclambda(\tau^-)=\veclambda(\tau^+)$, $\tilde H^*(\tau^-)=\tilde H^*(\tau^+)$ \cite{Hartl_survey_boundedstate} ( $\tau^-$ and $\tau^+$ denote the left and right limits, respectively).

\subsection{OCP for three-link swimmer}
We now formulate the displacement-per-cycle maximizing control problem for a three-link kinematic swimmer and find the solution. Whereas the formulation in \cite{wiezel2016using} used polar representation of the shape variables, here we write a more general formulation in order to allow for non-polar gaits.
For the three-link kinematic swimmer, we define the state as $\vecz=[\phi_1,\phi_2,\theta]^T$, with the dynamics:

\begin{equation}
\dot\vecz=\vcthree{\dot\phi_1}{\dot\phi_2}{\dot\theta}=\vcthree{u_1}{u_2}{p(\phi_1,\phi_2)u_1+q(\phi_1,\phi_2)u_2}
\label{eq:z_dot}
\end{equation}
with the input vector $\vecu=[u_1, u_2]^T=[\dot\phi_1,\dot\phi_2]^T$.
The cost function is the net-displacement over a period in the $x$-direction:
\begin{equation}
\resizebox{\linewidth}{!}{$J=x(t_f)=\int_0^{t_f} \dot x dt=\int_0^{t_f} g(\phi_1,\phi_2,\theta)u_1+h(\phi_1,\phi_2,\theta)u_2 dt$}
\end{equation}
The Hamiltonian $H$ is given by:
\begin{equation}
H(\vecz,\vecu,\veclambda)=gu_1+hu_2+\lambda_1u_1+\lambda_2u_2+\lambda_3(pu_1+qu_2)
\end{equation}
where $\veclambda$ is the vector of costate variables with dynamics:
\begin{equation}
\dot\veclambda=-\frac{\partial H}{\partial \vecz}=\resizebox{0.6\linewidth}{!}{$\vcthree{-g_{\phi_1}u_1-h_{\phi_1}u_2-\lambda_3(p_{\phi_1}u_1+q_{\phi_1}u_2)}{-g_{\phi_2}u_1-h_{\phi_2}u_2-\lambda_3(p_{\phi_2}u_1+q_{\phi_2}u_2)}{-g_{\theta}u_1-h_{\theta}u_2}$}
\label{eq:dot_lambda}
\end{equation}
with subscripts denoting partial derivatives, for example: $g_{\phi_1}=\frac{\partial g}{\partial \phi_1}$.
The Hamiltonian is linear in $u_1,u_2$, and the control inputs do not appear in the derivative:
\begin{equation}
\vecH_\vecu=\frac{\partial H}{\partial \vecu}=\vctwo{g+\lambda_1+\lambda_3p}{h+\lambda_2+\lambda_3q}
\label{eq:Hu}
\end{equation}
Since we may have $H_u=0$ for a finite time period, the optimal control following a singular arc may be found by requiring that the time derivatives $\dot\vecH_\vecu$ and $\ddot\vecH_\vecu$ vanish as well. Substituting \eqref{eq:z_dot} and \eqref{eq:dot_lambda} into the derivative of \eqref{eq:Hu} we have the first-order derivative as:
\begin{equation}
\dot \vecH_\vecu=\resizebox{0.75\linewidth}{!}{$\vctwo{u_2}{-u_1}\overbrace{(g_{\phi_2}-h_{\phi_1}+g_{\theta}q-h_{\theta}p+\lambda_3p_{\phi_1}-\lambda_3q_{\phi_1})}^{\Psi(\vecz,\veclambda)}$}
\label{eq:Hu_dot}
\end{equation}
Dismissing the trivial solution of zero control input, the requirement of $\ddot\vecH_\vecu=0$ reduces to the scalar equation $\frac{d}{dt}\Psi(\vecz,\veclambda)=0$, which leads to an equation in the form:

\begin{equation}
A(\vecphi,\theta,\lambda_3)u_1+B(\vecphi,\theta,\lambda_3)u_2=0
\label{eq:Hudd}
\end{equation}
The last equations gives the angle of the local tangent { to the gait's curve} $\frac{d\phi_2}{d\phi_1}$. Due to the time invariance property { of the motion, eq. \eqref{eq:geom_connection}, one only needs this tangent in order to construct the geometric shape of the optimal gait trajectory $\vecphi(s)$. Therefore, we arbitrarily choose time-scaling } $u_1^2+u_2^2=1$ and write the controls as $u_1=B/\sqrt{A^2+B^2},u_2=-A/\sqrt{A^2+B^2}$. The final time $t_f$ is unspecified.
Due to the symmetries of our system \cite{Gutman2015Symmetries}, we may restrict our analysis to one { quarter} of the gait (from $\phi_1=\phi_2$ to $\phi_1=-\phi_2$) and invoke symmetry. The total displacement will be four times the displacement over the quarter gait.
Because the rotation angle is known only at the initial time $\theta(0)=0$ and $\theta(t_f)$ is unknown, using transversality conditions \cite{bryson1975applied} we have $\lambda_3(t_f)=0$. We also have the boundary conditions on the shape variables $\phi_1(0)=\phi_2(0)$, $\phi_1(t_f)=-\phi_2(t_f)$.
The costate variables $\lambda_1,\lambda_2$ do not appear in the solution \eqref{eq:Hudd} and we do not need to solve for them.
Using the relation $\dot \vecH_\vecu =0$ we can find $\lambda_3(0)$ as a function of the state variables. We are left with a system of four ODEs for the variables $(\phi_1,\phi_2,\theta,\lambda_3)$ where all but $\phi_1$ are known at the initial time $t=0$. Using the shooting method we can find $\phi_1(0)$ that results in $\lambda_3(t_f)=0$ with the relation $\phi_1(t_f)=-\phi_2(t_f)$ defining the final time.

\subsubsection*{Solution with bounded joint angle}
Assume a practical bound on the joint angles $\left|\phi_i\right|\leq b$. Over a finite time { interval} with a non-zero control input, only the bound on one joint { angle } can be active, { while the other joint angle is varying}. Using the symmetry properties of the swimmer, we only consider one quadrant of the gait where only one joint may reach the bound. Therefore, it is sufficient to consider only a single, scalar state bound { of the form \eqref{eq:bound}}. For our demonstration, we will assume, without loss of generality, that the only bound is $\phi_2\leq b$.
We define
\begin{equation}
w(\vecz,t)=\phi_2-b\leq 0
\end{equation}
Assuming $\phi_2(0)<b$, the optimal gait starts on the singular arc and the control is found from \eqref{eq:Hudd}. On the bound $\phi_2=b$ we have $\dot w=u_2 =0$ which leads to $u_1=\pm 1$.
The entry to the bound $\tau_1$ is the time when the singular arc reaches the bound $\phi_2=b$. We determine the exit time $\tau_2$ using the shooting method to satisfy the end condition $\lambda_2(t_f)=\lambda_1(t_f)$. This means that we must solve for $\lambda_1,\lambda_2$ as well and ascertain the discontinuities at the entry and exit times.
As stated earlier, the discontinuity of the control inputs at the entry and exit point implies continuity of the costate variables and the Hamiltonian.
The transversality conditions lead to $\lambda_1(0)=-\lambda_2(0)$ and $\lambda_1(t_f)=\lambda_2(t_f)$. We can find $\lambda_1(0)$ from $\vecH_\vecu(0)=\vec{0}$.

\subsection{Maximum-displacement gaits for three-link swimmers}
We now present the solutions of the maximum displacement-per-cycle control problem for the three-link swimmer models presented above, and the influence of input bounds and swimmers' parameters on existence and topology of optimal solutions.
The method presented in the previous section results in a set of differential equations. These are solved using \textsc{Matlab}'s \texttt{ode45} function. Event functions detect crossing the bounds or reaching the final time at $\phi_1=-\phi_2$. When a bound is reached, the simulation continues along the bound until a exit time $\tau_2$. The initial value $\phi_1(0)=\phi_2(0)$ and the exit time $\tau_2$ are found using \texttt{fzero} function to satisfy $\lambda_1(t_f)=\lambda_2(t_f)$ and $\lambda_3(t_f)=0$

\subsubsection*{Purcell's swimmer}
For Purcell's three-link swimmer model, a gait for maximal displacement was presented in \cite{wiezel2016using} using a polar representation of the shape variables.
The optimal gait (seen in purple in Fig. \ref{fig:optimal_purcell_PMPcir}) is identical to that found numerically in \cite{tam2007optimal}.
Restricting to circular gaits, one can see (Fig. \ref{fig:Purcell_disp}) that there exists a second optimal amplitude, that when followed in the same direction (counter-clockwise) will give a net-displacement in the opposite direction with a greater absolute value. Hence, it is reasonable to expect a second, general, large amplitude, displacement-maximizing trajectory that would result in translation in the opposite direction.
Our attempts to find a second displacement-maximizing gait for Purcell's swimmer using \textit{PMP}   failed for the unbounded problem. Nevertheless, when applying bounds of $b=3.2$ [rad] to the problem, the gait shown in red in Fig. \ref{fig:optimal_purcell_PMPcir} is found. Gaits for some smaller bounds are also shown in Fig. \ref{fig:optimal_purcell_PMPcir}. Curiously, this method fails to find a solution for larger bounds.
In section \ref{sec:GAG} we attempt to use geometric analysis to explain this failure and why the unbounded problem does not have a solution.

\begin{figure}[t]
 \centering
 \begin{tikzpicture}
    \node[anchor=south west,inner sep=0] (image) at (0,0,0) {\includegraphics[width=0.3\textwidth]{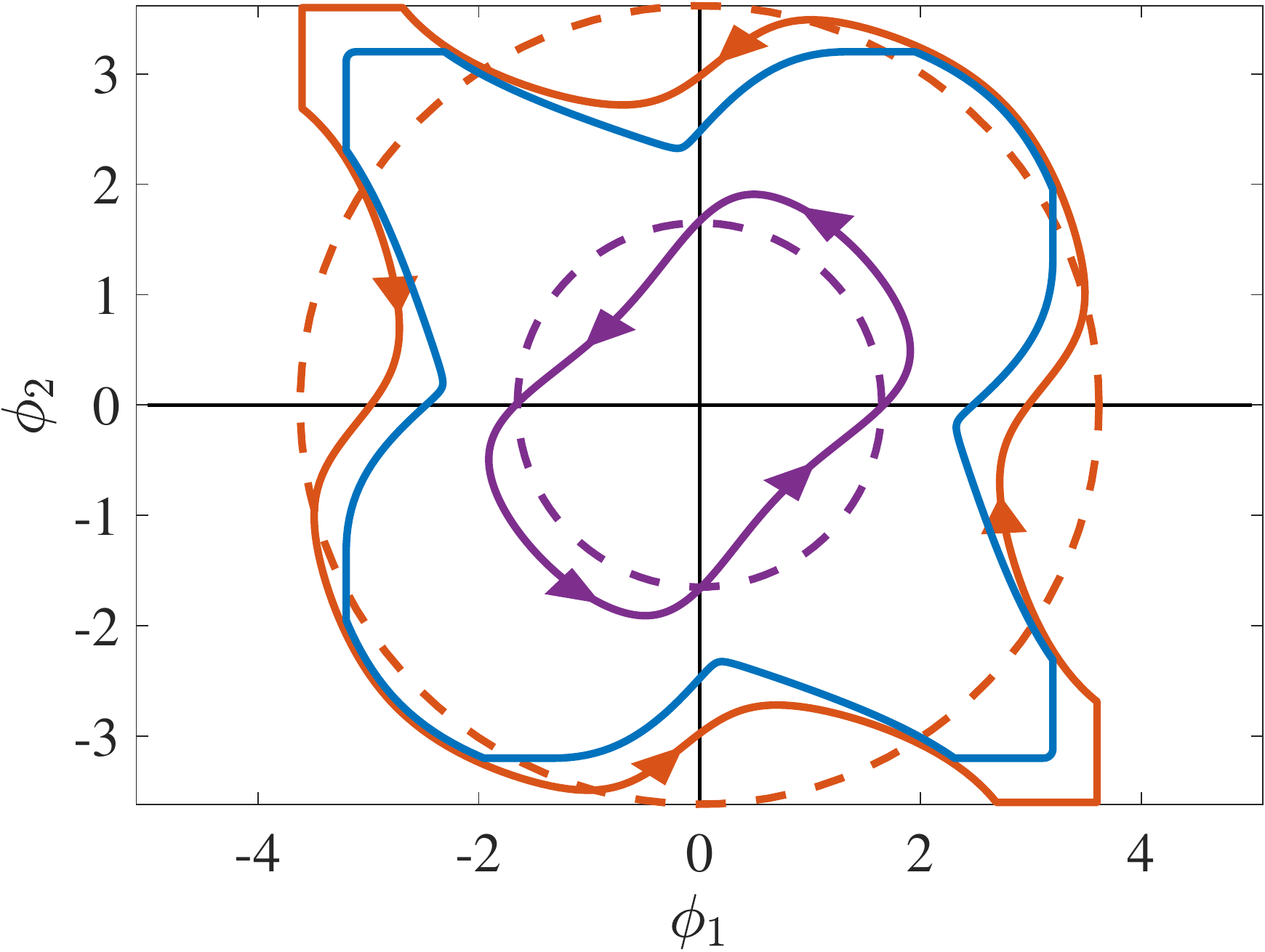}};
    \begin{scope}[x={(image.south east)},y={(image.north west)}]
        \draw [color=matred,thick](0.9,0.9) node{$X<0$};
        \draw [color=matpurple,thick](0.51,0.5) node{$X>0$};
    \end{scope}
\end{tikzpicture}
\caption{Maximal displacement per cycle gait (purple) and second optimal gait with bounds of $\phi_i\leq 3.6$ [rad] (red) and $\phi_i\leq 3.1$ [rad] (blue) for Purcell's swimmer. The dashed lines are the corresponding optimal circular gaits.}
 \label{fig:optimal_purcell_PMPcir}
 \end{figure}

 \begin{figure}[b]
\centering
\begin{subfigure}{0.24\textwidth}
\includegraphics[width=\textwidth]{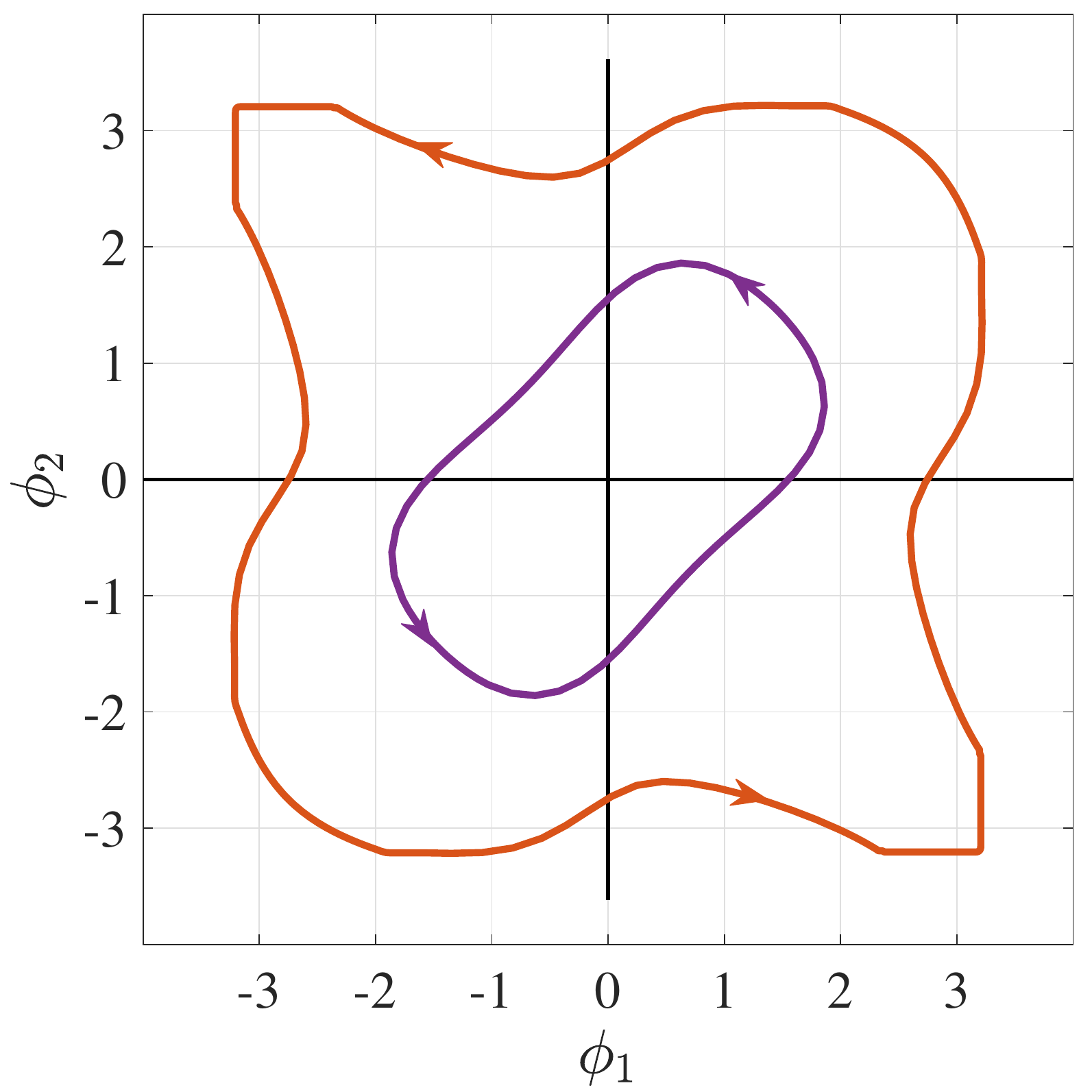}
\caption{}
\label{fig:perfect_eta_03_alpha_02}
\end{subfigure}
\begin{subfigure}{0.24\textwidth}
\includegraphics[width=\textwidth]{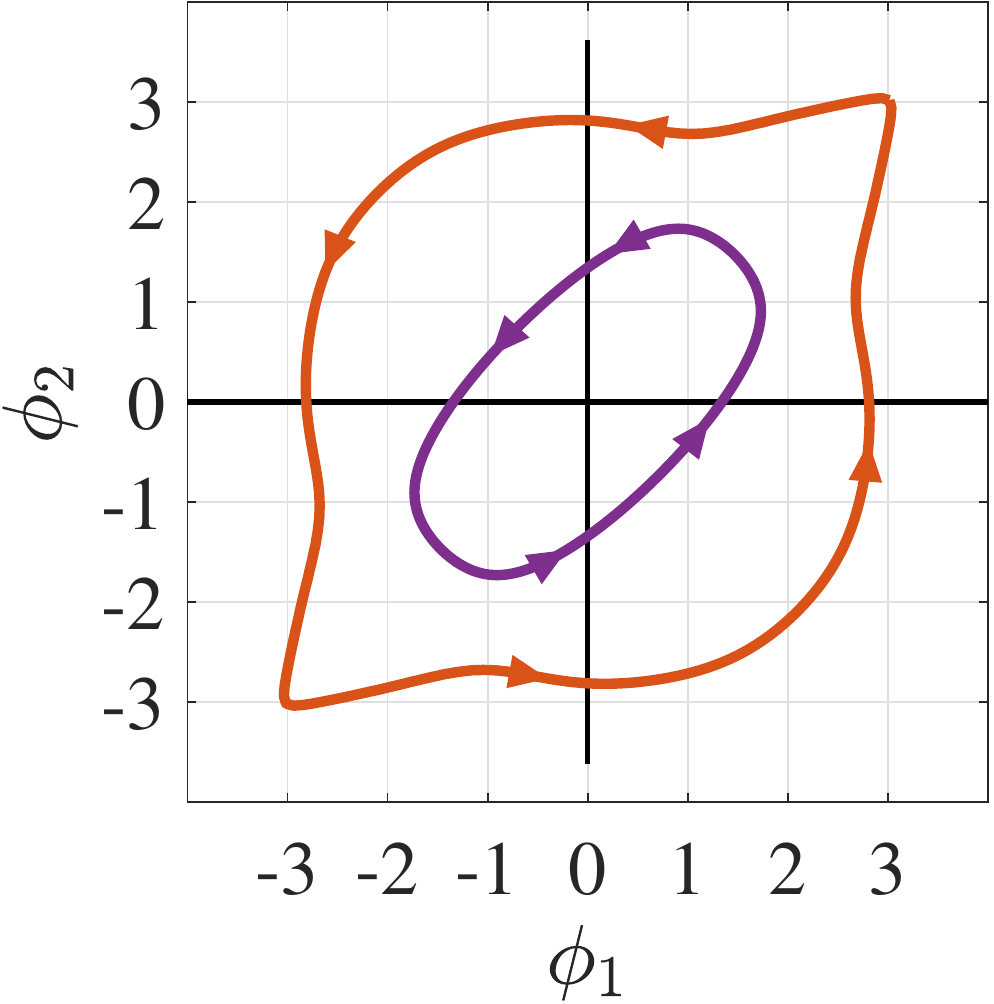}
\caption{}
\label{fig:optimal_perfect_PMP2}
\end{subfigure}
 \caption{Maximal displacement gaits for perfect fluid swimmer with (a) $\eta=1/3$ (red curve achieved only with bounded joints) and (b) $\eta=1/2$.}
 \label{Fig:perfect_PMP}
 \end{figure}

\subsubsection*{Perfect fluid swimmer}
The maximum-displacement-per-cycle gait for a perfect fluid swimmer model with elliptical links having radii ratio of $\alpha=a_i/b_i=0.2$ and links' length ratio $\eta=1/3$ is presented in Fig. \ref{fig:perfect_eta_03_alpha_02}. The gait is qualitatively similar to that found for Purcell's swimmer and can also be found by the polar formulation in \cite{wiezel2016using}. Here too, we can expect to find a second, large-angle gait that results in opposite displacement of the swimmer, $X<0$. However, for a swimmer with this specific geometry, the unbounded OCP had no solution for such gait.
Once again, when imposing bounds on the joint angles, a gait with $X<0$ is found for the perfect fluid swimmer as well. Fig. \ref{fig:perfect_eta_03_alpha_02} shows the second displacement-maximizing gait with joint angle bounds of $|\phi_i| \leq 3.2$.
Unlike Purcell's swimmer, for the perfect fluid model we found certain parameter values that would yield a second displacement-maximizing solution without imposing bounds. By changing the link length ratio to $\eta=1/2$, a second solution emerged to the unbounded problem, that corresponds to a second gait. Both gaits for the swimmer with $\eta=1/2$ are shown in Fig. \ref{fig:optimal_perfect_PMP2}.

Applying  \textit{PMP}  reveals  maximal displacement gaits for the swimmer models.
{ However, sometimes it fails to find an optimal solution, unless appropriate bounds on joint angles are incorporated, which enable finding optimal gait trajectories containing pieces along the bounds. }
In order to get a better understanding of the { influence of joint angle bounds on the existence of} 
displacement-maximizing gaits, in the following section we re-examine the kinematic swimmer models using a geometric approach.

\section{Geometric analysis of gaits}
\label{sec:GAG}
In order to have better insight into the behaviour of the kinematic models and understand the changes in the optimal gaits, we now review a geometric approach to the analysis of such swimmers \cite{hatton2010optimizing}. {This approach uses a generalization of Stokes' theorem to capture the relationship between the geometry of a gait's trajectory through shape space and the displacement it produces. For the purposes of this paper, a key result of the geometric approach is that the maximum-displacement-per-cycle gaits described in the previous section approximately follow the zero-contour of the \emph{curvature} (augmented curl) of the local connection, such that the shape of the zero contour can be used to predict the existence and form of the maximum-displacement-per-cycle gaits.}

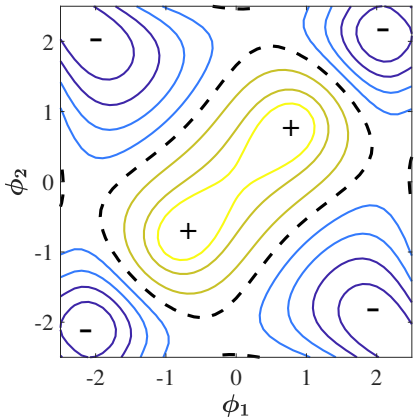
\begin{figure}[b]
 \centering
 \begin{tikzpicture}
    \node[anchor=south west,inner sep=0] (image) at (0,0,0) {\includegraphics[width=0.3\textwidth]{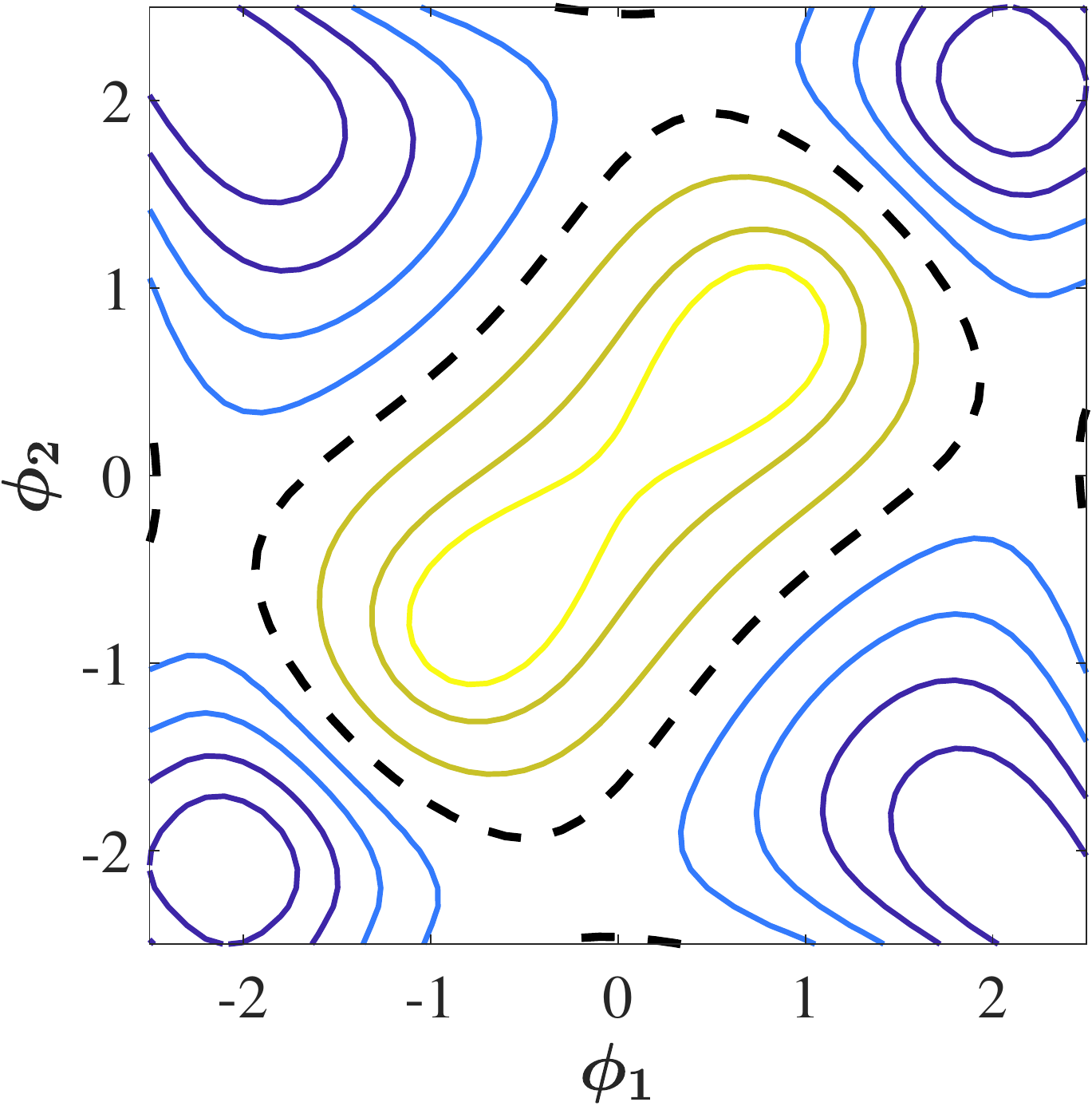}};
    \begin{scope}[x={(image.south east)},y={(image.north west)}]
        \draw (0.7,0.7) node{\large +};
        \draw (0.45,0.45) node{\large +};
        \draw (0.225,0.9) node{ \LARGE -};
        \draw (0.925,0.925) node{\LARGE -};
        \draw (0.9,0.25) node{\LARGE -};
        \draw (0.2,0.2) node{\LARGE -};
    \end{scope}
\end{tikzpicture}
\caption{The height function plot for Purcell's swimmer. Positive regions are in shades of blue and negative regions in shades of yellow. The zero level curve is the dashed black line.}
 \label{Fig:heightpurcell}
 \end{figure}

\subsection{Using Stokes' Theorem for Measuring Net Displacement}

{As a preliminary to discussing the geometric locomotion formulation, } we first give a brief review of Stokes' theorem. Let $S$ be an oriented smooth surface that is bounded by a simple, closed, smooth boundary curve $\Gamma$ with positive orientation. Also, let $\vecF$ be a vector field. Stokes' theorem states that the line integral along the closed curve $\Gamma$ on the vector field $\vecF$ is equal to the integral of the curl of that vector field over a surface bounded by the curve,
\begin{equation}
\ointctrclockwise_\Gamma \vecF\cdot d\vecr=\iint_S \text{curl}\,\vecF\cdot d\vecS
\end{equation}
{Intuitively, $\text{curl}\,\vecF$ measures how much $\vecF$ changes ``across" the gait, such that the contribution from flowing along one side of the gait is not undone by flowing back along the opposite side of the gait.}

Next, we show how the same logic as underlies Stokes' theorem can be applied in order to approximate the net displacement of the swimmer. The dynamic equations of motion for the three-link swimmers, as given in \eqref{eq:full_connection}, are
\begin{equation}
\dot \vecq=\vecG\left(\vecphi,\theta\right)\vecphid = \vecR(\theta)\vecA\left(\vecphi\right)\vecphid
\label{eq:EOM3link}
\end{equation}
The net displacement of the swimmer over a period is equal to the line integral over the gait,
\begin{equation}
\Delta \vecq =
\ointctrclockwise_{\vecphi}\vecR(\theta)\vecA\left(\vecphi\right)
\label{eq:line_int}
\end{equation}

This line integral can be approximately converted to a surface integral in a manner similar to Stokes' theorem by evaluating the {curvature of the local connection} over the surface $\vecphi_a$ as in \cite{ramasamy2016soap}, giving the ``corrected Body Velocity Integral'' {(cBVI)

\begin{equation}
    \Delta \vecq \approx \exp{\iint_{\vecphi_a}D\vecA} = \exp{\iint_{\vecphi_a}\mathrm{d}\vecA-[\vecA_1,\vecA_2]}
    \label{eq:surfint}
\end{equation}
}
Here, $\mathrm{d}\vecA$ is the exterior derivative of the local connection (the generalized row-wise curl), and $[\vecA_1,\vecA_2]$ is a local Lie bracket term that corrects for noncommutativity as the swimmer translates and rotates through space {(e.g. if the system rotates counterclockwise before moving forward and/or clockwise before moving backward, then the influence of the $\vecR(\theta)$ term means that the ``forward" and ``backward" motions will be tilted slightly ``leftward".) This bias is captured to first order by the Lie bracket, which evaluates for planar translation and rotation as}
\begin{equation}
    [\vecA_1,\vecA_2] = \begin{bmatrix}\vecA_1^y\vecA_2^{\theta} - \vecA_2^y\vecA_1^{\theta} \\
    \vecA_2^x\vecA_1^{\theta} - \vecA_1^x\vecA_2^{\theta} \\ 0
    \end{bmatrix}
\end{equation}
{The $\exp$ operator in \eqref{eq:surfint} converts the cBVI from an ``average body velocity" to a displacement by flowing with the cBVI as body velocity for unit time. If the rotational component of the cBVI is zero, the components of $\exp{\text{cBVI}}$ are the same as those of the $\text{cBVI}$; if the rotational component is non-zero, then the exponential flow is curved commensurately, approximating the net rotation as having been evenly distributed along the system's motion.}


\subsection{Height Functions and Displacement-Per-Cycle Maximizing Gaits}


Using the generalized Stokes' theorem , we have a good approximation of the net displacement of the swimmer over a gait. By plotting the x-integrand in \eqref{eq:surfint} as a height function $H_x\left(\vecphi\right) = D\vecA^x(\vecphi)$, we can identify sign-definite areas of the shape plane. A positive area that is encircled by the gait (in a counter-clockwise direction) will result in positive net displacement of the swimmer in the $x$ direction. In Fig. \ref{Fig:heightpurcell} a contour plot of the height function for the three-link swimmer is presented.
In order to maximize the displacement of the swimmer, the gait should enclose a region of the height function that is as sign-definite as possible. Obviously, this will be accomplished by following the \textit{zero level curve} that separates the positive and negative regions. The zero level curve for the symmetric three-link swimmer, represented in Fig. \ref{Fig:heightpurcell} by a dashed closed curve, is functionally identical to the optimal gait found using \textit{PMP}   in \cite{wiezel2016using} and numerically in \cite{tam2007optimal}.

{
\subsection{Minimum Perturbation Body Frames}
Note that our geometric approach involves two approximations for handling the noncommutativity induced by the $\vecR(\theta)$ term: the Lie bracket that makes a first-order approximation of the results of mixing \emph{intermediate} translation and rotation within a cycle by identifying whether forward/backward motion was on-average preceded by clockwise/counterclockwise motion;  and the exponential map that takes the \emph{net} rotation over a cycle as having been evenly distributed over the course of the motion. If these approximations are not accurate, then the approximated net displacement will have some residual error.}


{A core finding from our prior work~\cite{hatton2011geometric,hatton2015epj,bass2022characterizing} is that the choice of body frame coordinates affects the magnitude of this residual error, and in particular that the error correlates to the norm of the local connection $\vecA$. For example, body frames attached to outer links have large $\vecA$ values and so experience large motions in response to shape changes. These motions accrue considerable error, both because they rotate outside of the $\sin(\theta) \approx \theta$ linear approximation for rotation and because they have significant translations while in these heavily-rotated configurations. Conversely, frames such as those attached to the center of mass and oriented to align with the mean of the link angles have small $\vecA$ values, and this experience small motions that stay within the valid domains of the approximations. (Note that the ``large" and small" motions under mentioned here are the intermediate motions; the net rigid body motion is independent of the choice of body frame.)

{For the examples in this paper, we placed the body frame at a location $\bar{\vecq}_b=(\bar x ,\bar y , \bar \theta )$ found through our algorithm in~\cite{hatton2011geometric,hatton2015epj}, which identifies a weighted average of the link positions and orientations that variationally minimizes the average norm of $\vecA$ over the region of the shape space under consideration.}

} 

\section{Explaining changes in optimal gaits using geometric analysis}
\label{sec:compare}


After reviewing the geometric method for gait analysis, we can now revisit the cases where optimal gaits derived by \textit{PMP}   analysis depend on varying the swimmer's parameters and/or state bounds. By using the minimum perturbation coordinates and plotting the height function for the models discussed here, we get a better understanding of the way changes in the gait affect the displacement. Moreover, we can see that topological changes in the zero-level curves induce formation of junction points for which the variational equations underlying \textit{PMP}   method become singular.

\subsection{``Reverse optimum'' for Purcell's swimmer.}

First, we revisit the case of the symmetric three-link swimmer and show a contour plot of the height function $H_x$ under the minimum-perturbation coordinates. Fig. \ref{fig:perfect_03_height} shows the height function for the perfect fluid swimmer at larger amplitudes of up to $6$ [rad]. In this plot, an additional zero level curve can be seen in amplitudes of around $3$ [rad]. The junctions, which can clearly be seen on the curve, explain the failure of the \textit{PMP}   method to find an optimal gait. These junctions mean that there is no unique solution to the unbounded optimal control problem presented.
\subsection{The perfect fluid swimmer.}
For the perfect fluid model we found two gaits with opposite displacement directions for a link length ratio of $\eta=1/2$, but were unable to find a second, reverse direction, optimal gait for a ratio of $\eta=1/3$ without applying bounds to the joint angles. Height functions for both cases are presented in Fig. \ref{Fig:perfect_PMP_vs_height}. In Fig. \ref{fig:perfect_05_height}, showing the height function for $\eta=1/2$, the two zero-level curves can be seen in black { dashed lines}. These curves correspond to the two optimal gaits found using \textit{PMP}   (Fig. \ref{fig:optimal_perfect_PMP2}). For $\eta=1/3$, the height function in Fig. \ref{fig:perfect_03_height} shows the first zero level curve representing the optimal gait which was found using \textit{PMP}   (Fig. \ref{fig:perfect_eta_03_alpha_02}), while the second zero level curve has junctions, once again explaining the failure of \textit{PMP}   to find the second optimal gait for the unbounded problem.

Using the geometric method as presented here to plot the height function for the displacement of a swimmer gives insight on the optimal gait and offers an explanation for the cases where the \textit{PMP}   method fails to find an optimal gait, as well as topological changes in the optimal gaits.

\begin{figure}[h]
\centering
\begin{subfigure}{0.23\textwidth}
\includegraphics[width=\textwidth]{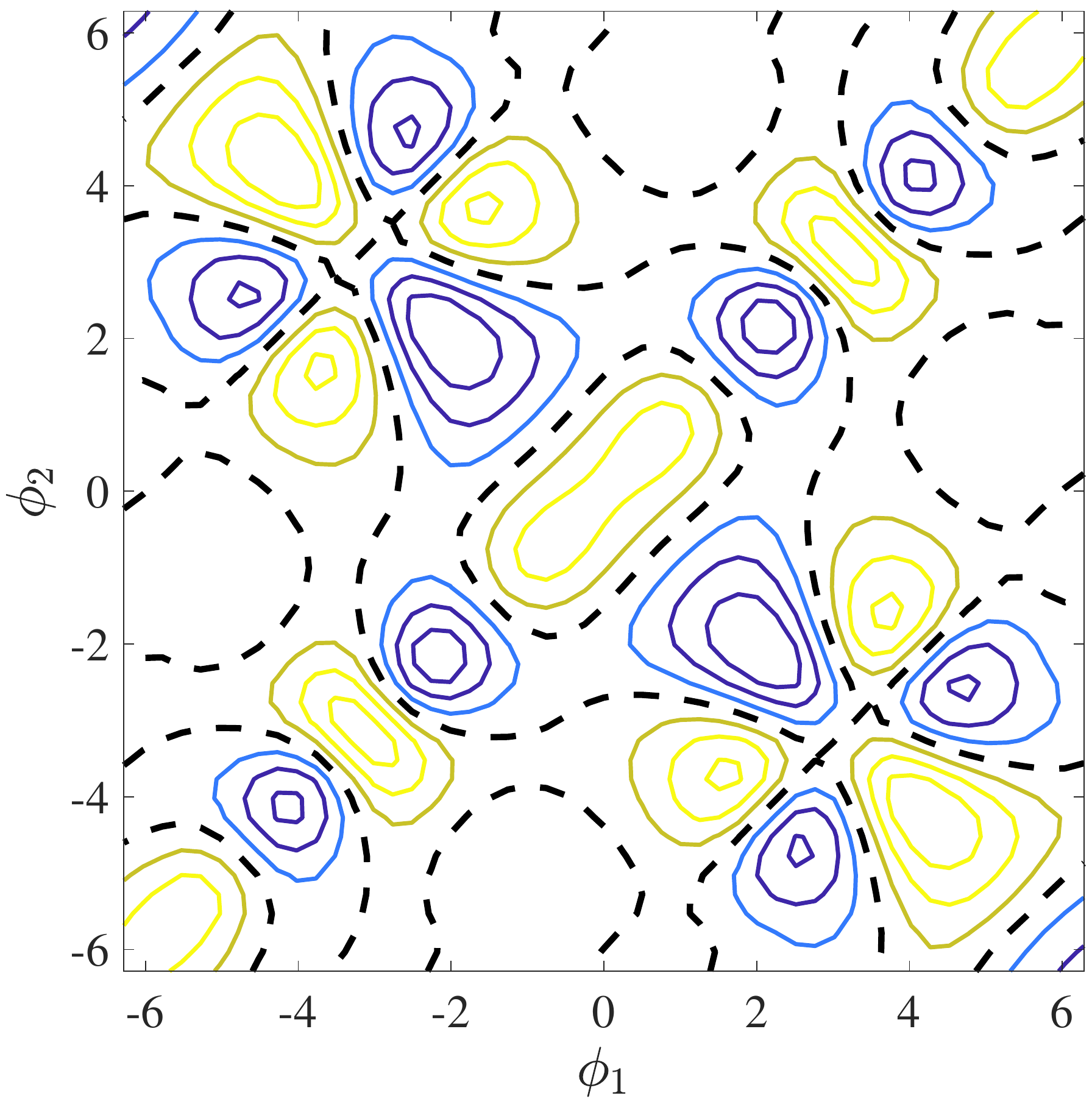}
\caption{}
\label{fig:perfect_03_height}
\end{subfigure}
\begin{subfigure}{0.23\textwidth}
\includegraphics[width=\textwidth]{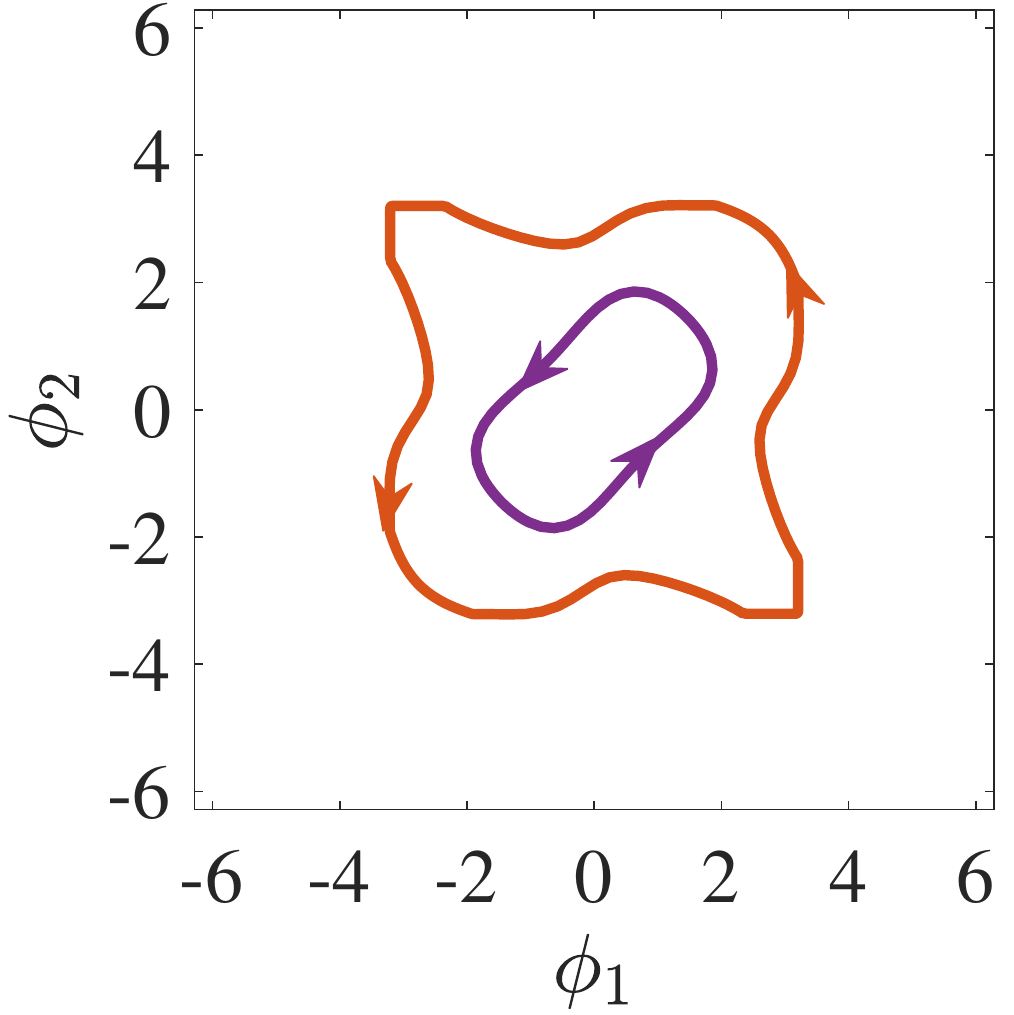}
\caption{}
\label{fig:perfect_eta_03_alpha_02second}
\end{subfigure}
\begin{subfigure}{0.23\textwidth}
\includegraphics[width=\textwidth]{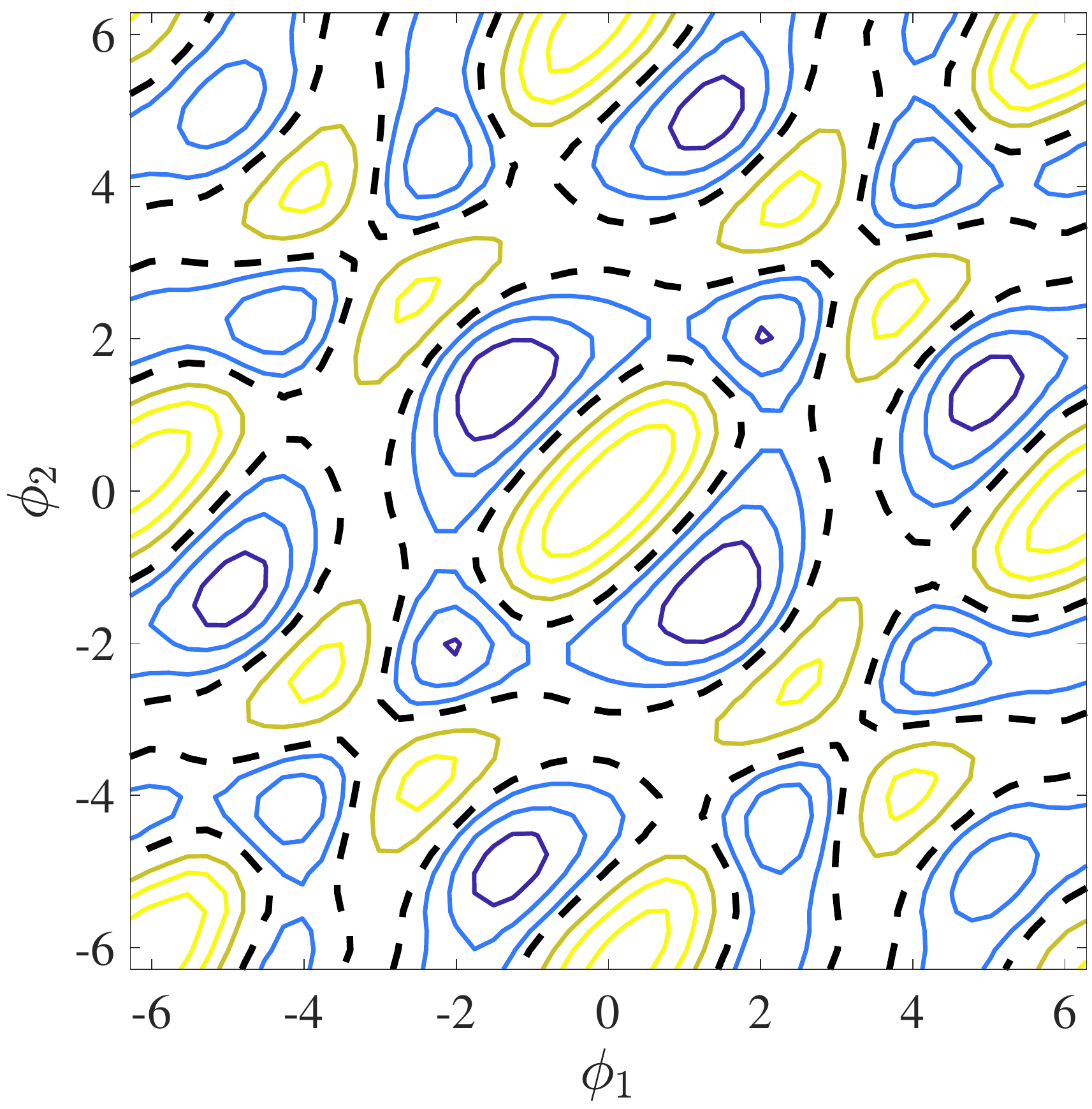}
\caption{}
\label{fig:perfect_05_height}
\end{subfigure}
\begin{subfigure}{0.23\textwidth}
\includegraphics[width=\textwidth]{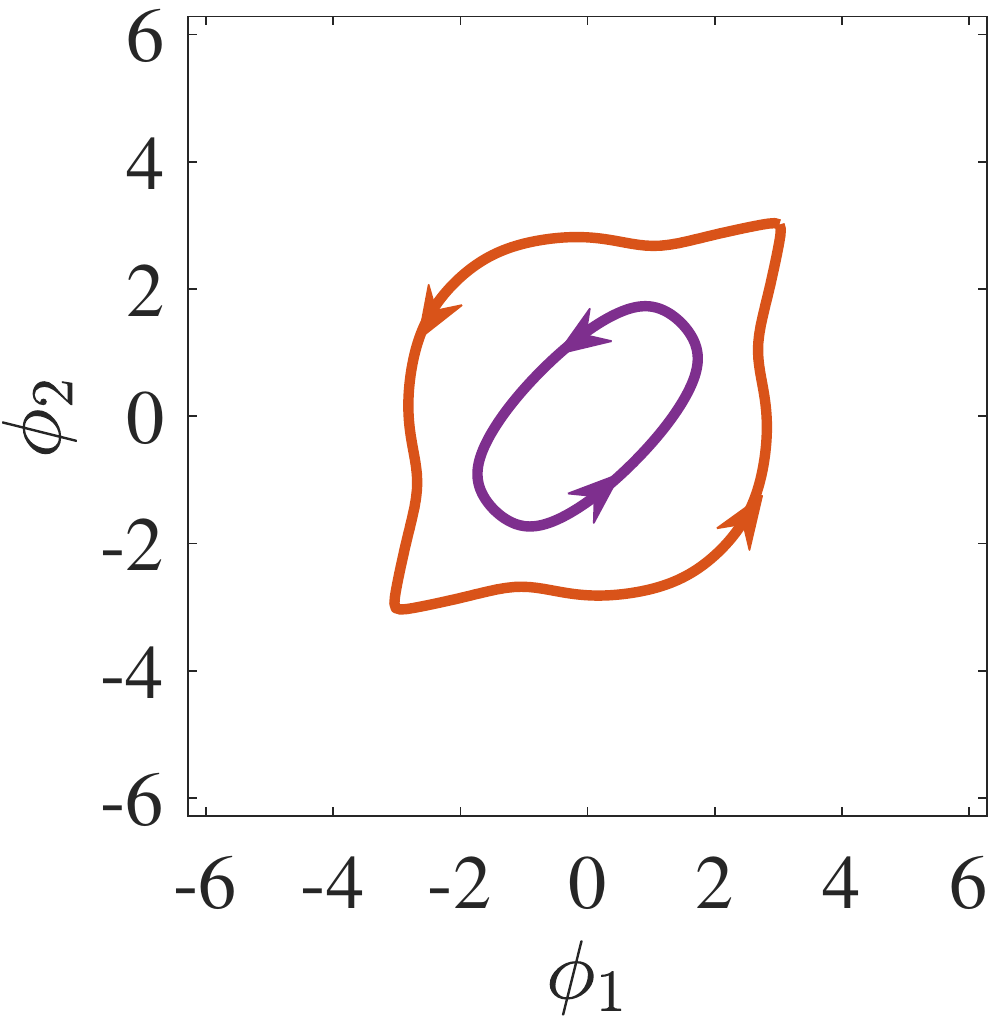}
\caption{}
\label{fig:optimal_perfect_PMP2second}
\end{subfigure}
 \caption{Height functions for ``perfect fluid'' swimmer with $\alpha=0.2$ (a) $\eta=1/3$. Junctions can be seen in the larger zero level curves {(dashed black lines)}. (b) The corresponding \textit{PMP}   solutions. (c) $\eta=1/2$. Two { loops of} zero level curves can be seen, corresponding to the two optimal gaits found via \textit{PMP}  . (d) Corresponding \textit{PMP}   solutions.}
 \label{Fig:perfect_PMP_vs_height}
 \end{figure}

\section{conclusion}
\label{sec:conc}
In this paper we examined and compared different methods of finding displacement-maximizing gaits for two models of 3-link swimmers.
We presented the two swimmer models: Purcell's three-link swimmer, and the `perfect fluid' swimmer. We formulated the optimal control problem for a general case of a three link swimmer with two joint angle inputs. We presented the solution using \textit{PMP}   and also considered the case where there are bounds on the joint angles.
In some cases, we found that the \textit{PMP}   method fails to find the maximum displacement gait for unbounded joint angles.
We then reviewed the geometric analysis approach to finding such gaits.
Observing the results from this method offered an explanation for the failures of \textit{PMP}   and to the unexplained changes in the optimal gait.
Junctions seen in the zero level curve mean there is no unique solution to the optimal control problem, which causes \textit{PMP}   to diverge.
Adding bounds to the problem allowed us to avoid these junctions.
The geometric method gives insight into the changes in the distance-optimal gait due to changes in the swimmer model and parameters.
We now briefly discuss limitations of our work and sketch some possible directions for future extensions of the research.
First, the methods used in this paper for displacement-maximizing gaits can be applied to finding energy-optimal gaits as well \cite{ramasamy2019geometry,tam2007optimal}.
Second, the work may be extended to time-dependent swimmer models, such as a swimmer with elastic joints \cite{passov2012dynamics} and a magnetically actuated swimmer \cite{gutman2014simple}, {\cite{zoppello2018modeling,el2020optimal}}.
Swimmers with multiple joints are another avenue of research { \cite{giraldi2013controllability,marchello2022n}}, \cite{Ramasamy2017GeometricBeyond}. In \cite{alouges2019energy} energy-optimal gaits were found for an $N$-link swimmer assuming small amplitudes of the joint angles. This could possibly be extended to large amplitudes by applying \textit{PMP}   method.
Finally, the results found here can be applied to real robotic micro swimmers in order to verify the validity of the results \cite{jang2015undulatory,HelicalOrNelsonChapnik2021,el2020optimal}.

{{\bf Acknowledgement:} The authors wish to thank the anonymous reviewers for their useful and important suggestions. YO also thanks Prof. Joseph Ben-Asher from Technion for guidance and fruitful discussions.}

\bibliographystyle{IEEEtran}
\bibliography{mybibliography}
\end{document}